# Data-driven multinomial random forest: A new random forest variant with strong consistency

JunHao Chen[1*], XueLi Wang[1], Fei Lei[2]

[1]Faculty of Science, Beijing University of Technology, Beijing, China
[2]Faculty of Information Technology, Beijing University of Technology, Beijing, China

## Abstract

In this paper, we modify the proof methods of some previously weakly consistent variants of random forests into strongly consistent proof methods, and improve the data utilization of these variants in order to obtain better theoretical properties and experimental performance. In addition, we propose a data-driven multinomial random forest (DMRF), which has the same complexity with BreimanRF (proposed by Breiman) while satisfying strong consistency with probability 1. It has better performance in classification and regression problems than previous RF variants that only satisfy weak consistency, and in most cases even surpasses BreimanRF in classification tasks. To the best of our knowledge, DMRF is currently a low-complexity and high-performing variation of random forests that achieves strong consistency with probability 1.

## 1. Introduction

Random Forest (RF, also called standard RF or BreimanRF)[1] is an ensemble learning algorithm that makes classification or regression predictions by taking the majority vote or average of the results of multiple decision trees. Due to its simple and easy-to-understand nature, rapid training, and good performance, it is widely used in many fields, such as data mining[2,3,4], computer vision[5,6,7], ecology[8,9], and bioinformatics[10].

Although the RF algorithm has excellent performance in practical problems, analyzing its theoretical properties is quite difficult due to its highly data-dependent tree-building process. These theoretical properties include consistency, which can be weak or strong. Weak consistency refers to the expectation of the algorithm's loss function converges to the minimum value as the data size tends to infinity, while strong consistency refers to the algorithm's loss function itself converges to the minimum value as the data size tends to infinity[11]. Consistency is an important criterion for evaluating whether an algorithm is excellent, especially in the era of big data.

[*] Corresponding author. E-mail address: chenjunhao@emails.bjut.edu.cn.

Many researchers have made important contributions to the discussion of consistency-related issues in RF, proposing many variants of RF with weak consistency, such as Denil14 (also called Poisson RF)[12], Bernoulli RF (BRF)[13], and Multinomial RF (MRF)[14]. However, the common feature of these algorithms is that the selection of split points and the determination of the final leaf node labels during the tree-building process are independent, i.e., using half of the training set samples to train the split points and the remaining half to determine the leaf node labels, which to a large extent causes insufficient growth of the basic decision trees. In addition, the introduction of random distributions such as Poisson distribution, Bernoulli distribution, and multinomial distribution in these algorithms can enhance their robustness but also have a certain impact on their performance.

In this paper, we propose a new variant of random forest algorithm called the Data-driven Multinomial Random Forest (DMRF) which has strong consistency, based on the foundation of MRF and BRF with weak consistency. The term "Data-driven" in this context does not mean that other variants of random forest algorithm are not data-driven, but rather indicates that the DMRF algorithm can make more effective use of data compared to the aforementioned variants with weak consistency. In the DMRF algorithm, we incorporate a new bootstrapping (slightly different from the standard bootstrapping in BreimanRF) that was not included in the previous variants, and introduce a Bernoulli distribution when splitting nodes. This Bernoulli distribution determines whether using optimal splitting criterion to obtain the splitting point (a splitting point consists of a splitting feature and a splitting value), or sampling the splitting point using two multinomial distributions based on impurity reduction. The reason for introducing the multinomial distribution is that it can perform random sampling of the optimal splitting feature and feature value with maximum probability[14].

## 2. Related work

BreimanRF[1] is an ensemble algorithm based on the prediction results of multiple decision trees, proposed by Breiman. It has shown satisfactory performance in practical applications. The basic process of BreimanRF can be divided into three steps: first, use bootstrapping to resample the dataset for the same number of times as the size of the dataset to obtain the training set for the basic decision tree; second, randomly sample a feature subspace of size $\sqrt{D}$ without replacement from the entire feature space of size $D$, and evaluate the importance of each feature and feature value in the subspace based on the reduction in impurity (e.g., Information entropy or Gini index) to obtain the optimal splitting point. Recursively repeat this process until the stopping condition is met, and a decision tree is obtained. Finally, repeat the above process to train multiple decision trees, and take the majority vote (for classification problems) or average (for regression problems) of the results of multiple decision trees to obtain the final prediction result.

Since the proposal of the RF model, many variants have been developed, such as Rotation Forest[15], Quantile RF[16], Skewed RF[17], Deep Forest[18], and Neural RF[19]. These variants are proposed to further enhance the interpretability, performance, and efficiency of the random forest. Although the practical research on RF has developed rapidly, the progress of its theoretical exploration is slightly lagging behind. Breiman proved that the performance of BreimanRF is jointly determined by the correlation between basic trees and the performance of basic trees, i.e., the smaller the correlation between trees (i.e., the greater the diversity), and stronger the trees (i.e., the better the performance), the better the performance of RF[1].

The important breakthrough in the study of consistency of RF was proposed by Biau et al.[20]. Biau proposed two simplified algorithms of BreimanRF: Purely Random Forest and Scale-Invariant Random Forest. Purely Random Forest randomly selects a feature and its feature value as the splitting feature and splitting value at each node. Scale-Invariant Random Forest also randomly selects a feature as the splitting feature at each node and randomly divides the samples into two parts according to the order of the feature values of that feature. Biau proved that both of these simplified versions have consistency.

Biau[21] proved another simplified RF model that is closer to BreimanRF and has weak consistency: randomly selecting a feature subspace at each node, and for each candidate feature, selecting the midpoint as the splitting value. When selecting among candidate features, the splitting feature and splitting value with the maximum reduction in impurity are selected to grow the tree.

Denil et al.[12] proposed a new RF variant, Denil14, which is closer to RF.

Denil14 divide the training set into structural part and estimation part. The structural part is used only for training the split points, and the estimation part is used only to determine the labels of the leaf nodes. In addition, at each node, the size of the feature subspace is selected based on the Poisson distribution, and the optimal splitting feature and splitting value are searched from $m$ structural part samples that have been pre-selected. This variant has been proven to have weak consistency. Denil14 can be used for classification.

Inspired by the Denil14 model, Yisen Wang et al.[13] proposed a new Bernoulli RF (BRF) based on the Bernoulli distribution. Similar to Denil14, BRF divides the dataset into structural part and estimation part, trains the splitting points using the structural part, and determines the leaf node labels using the estimation part. However, BRF introduces two Bernoulli distributions at each node: one to determine whether the feature subspace size is 1 or $\sqrt{D}$, and the other to determine whether the splitting value of each candidate feature is randomly selected or using the optimal splitting criterion. Finally, they proved that BRF also has weak consistency, but it is closer to BreimanRF and has better performance than previous RF variants with weak consistency. BRF can be used for both classification and regression.

Jiawang Bai et al.[14] transformed the reduction of each feature and its impurity value into probabilities through the softmax function, proposing a Multinomial Random Forest (MRF). MRF also divides the dataset into structural and estimation parts, using the structural part to train the splitting point and the estimation part to determine the leaf node label. Training the splitting point primarily involves two steps: first, calculating the maximum impurity reduction for each feature at each node and converting it into a probability, which is then considered the probability of the multinomial distribution from which the splitting feature is randomly selected; second, when selecting the splitting value, converting the impurity reduction of each feature value of the split feature obtained in the previous step into a probability and regarding it as the probability of the multinomial distribution from which the splitting value is randomly selected. Regarding determining the leaf node label, MRF views the proportion of each class of estimation part samples in the leaf node as a probability and randomly selects a class as the label from the multinomial distribution. MRF has more purposeful random selection of splitting features and values, with more reasonable probability allocations. It currently has the best performance among the consistent variants of RF, even surpassing BreimanRF. The disadvantage is its high computational complexity, it can cause large computational cost. MRF is only used for classification.

Due to the fact that Denil14, BRF, and MRF all chose to make the training of split points and leaf node labels independent to achieve weak consistency, this inevitably affects the performance of the base tree and reduces the overall performance of the algorithm. Based on this, we propose the Data-driven Multinomial Random Forest (DMRF) algorithm that can be used for both classification and regression problems. This algorithm can directly select the sample for the training of the splitting point to determine the leaf node label. Moreover, this algorithm introduces bootstrapping to increase the diversity between trees. More importantly, we enhance the weak consistency to strong consistency by modifying the conditions of the variants mentioned above. We found that although the theoretical basis is different, the proof methods are quite same as before. This means that the method for proving weak consistency can be strengthened to the method for proving strong consistency, resulting in a strong conclusion.

The arrangement of this paper is approximately as follows: Section 3 provides a detailed introduction to the classification and regression DMRF algorithm and proves its strong consistency; Section 4 provides some explanations of the experiments; Section 5 presents the experimental results and analysis; Section 6 concludes the paper and provides a future outlook on the work.

## 3. The proposed DMRF algorithm

### 3.1 Classification DMRF

Let $D_n = \{(X_1, Y_1), (X_2, Y_2), ..., (X_n, Y_n)\}$ denotes a dataset where $X_i \in \mathcal{R}^D$ indicates $D$-dimensional features, $Y_i \in \{1, 2, ..., c\}$ indicates the label, $i \in \{1, 2, ..., n\}$, we are preparing to build $M$ trees.

### 3.1.1 Training sample sampling

In the DMRF, we use a slightly different bootstrapping than the standard one to sample the training set $D_n^{(j)}$, $j \in \{1, 2, ..., M\}$ for the $j$-th tree. Specifically, during sampling, we do not resample all samples, but instead sample each sample with a probability of $q$ (which may be related to $n$, in this paper, we choose a constant value in $(0,1]$). That is, the sampling of each sample follows a binomial distribution $B(1, q)$, and the probabilities $q$ among different samples are independent. If the sampled training set is empty, we resample again.

### 3.1.2 Split point training process

First, let's introduce a very important function: the softmax function.

**Definition 3.1**: Given a $n$-dimensional vector $v = (v_1, v_2, ..., v_n)$, the softmax function is defined as follows:

$$\text{soft}\max(v) = (e^{v_1}, e^{v_2}, ..., e^{v_n}) / \sum_{i=1}^{n} e^{v_i},$$

where $e$ is the base of the natural logarithm. Obviously, after the transformation by softmax function, the elements of the vector are all numbers between 0 and 1, and their sum is equal to 1. Therefore, they can be seen as probabilities.

Reviewing the growth process of a classification tree in BreimanRF: In the node $\mathcal{D}$, $\sqrt{D}$ features are randomly selected to form a feature subspace, which denote as $\{A_1, A_2, ..., A_{\sqrt{D}}\}$ (Without loss of generality, we assume that $\sqrt{D}$ is an integer. Otherwise, we can round it down to the nearest integer). These selected features are also referred to as candidate split features. Let $V = \{v_{ij}\}$ denotes all possible split points for the node $\mathcal{D}$, $v_{ij}$ representing the $A_j$'s $i$-th feature value (i.e., threshold) in the feature subspace, $i \in \{1, 2, ..., m_j\}$, $j \in \{1, 2, ..., \sqrt{D}\}$, $m_j$ indicates the number of feature values for $A_j$. Let $I_{ij}$ denote the impurity reduction obtained by the split point $v_{ij}$ at that node,

$$I_{ij} = I(\mathcal{D}, v_{ij}) = T(\mathcal{D}) - \frac{|\mathcal{D}^l|}{|\mathcal{D}|} T(\mathcal{D}^l) - \frac{|\mathcal{D}^r|}{|\mathcal{D}|} T(\mathcal{D}^r), \qquad (1)$$

where $T(\mathcal{D})$ denote impurity criteria of node $\mathcal{D}$, such as Gini index or Information entropy (in this paper, we use the Gini index), are used for measuring impurity, and $\mathcal{D}^l$ and $\mathcal{D}^r$ respectively represent the left and right child nodes obtained by the split at that node.

The impurity reduction obtained by calculating different feature values of feature $A_j$ as split values forms a vector $I^{(j)} = (I_{1,j}, I_{2,j}, ..., I_{m_j, j})$, $j \in \{1, 2, ..., \sqrt{D}\}$.

The maximum impurity reduction for each feature forms a vector

$$I = (I_1, I_2, ..., I_{\sqrt{D}}) = (\max I^{(1)}, \max I^{(2)}, ..., \max I^{(\sqrt{D})}).$$

For a classification tree in BreimanRF, the splitting point is determined by the feature and feature value corresponding to the maximum impurity reduction, i.e., the splitting feature is the $j$-th feature, $j = \arg\max\{I_1, I_2, ..., I_{\sqrt{D}}\}$ and the splitting feature value is the $i$-th feature value of feature $A_j$, $i = \arg\max I^{(j)}$. After splitting the root node into left and right child nodes, the above process is repeated in each child node until the stopping condition is met, the tree stops growing.

The construction of DMRF tree is different from the above tree construction process. The following parameters will be introduced: $p$ is a probability, $k_n$ is the minimum sample number in a node, $B_1$, $B_2$ are two positive finite parameters.

DMRF first randomly samples $\sqrt{D}$ features from full feature space, then conducts a Bernoulli experiment $B$ with probability $p$ when splitting nodes, $B \sim B(1, p)$:

If $B = 1$, the split feature and split value at this node are obtained according to the optimal split criterion.

If $B = 0$, the split feature and split value at this node can be obtained by the following steps:

① Split feature selection:

    i) Normalize $I = (I_1, I_2, ..., I_{\sqrt{D}}) = (\max I^{(1)}, \max I^{(2)}, ..., \max I^{(\sqrt{D})})$ as

$$\tilde{I} = (I_1 - \min I, I_2 - \min I, ..., I_{\sqrt{D}} - \min I) / (\max I - \min I);$$

    ii) Compute the probablities $\alpha = soft\max(B_1 \tilde{I})$, where $B_1 \geq 0$;

    iii) Randomly select a splitting feature according to the multinomial distribution $M(\alpha)$.

② Split feature selection:

Assuming $A_j$ is the split feature which is selected from the previous step.

    i) Normalize $I^{(j)} = (I_{1,j}, I_{2,j}, ..., I_{m_j, j})$ as

$$\tilde{I}^{(j)} = (I_{1,j} - \min I^{(j)}, I_{2,j} - \min I^{(j)}, ..., I_{m_j,j} - \min I^{(j)}) / (\max I^{(j)} - \min I^{(j)}) ;$$

ii) Compute the probablities $\beta = soft\max(B_2 \tilde{I}^{(j)})$, where $B_2 \geq 0$;

iii) Randomly select a splitting value according to the multinomial distribution $M(\beta)$.

Continue repeating the above steps until a stopping condition is met, i.e. the number of samples within a node is less than $k_n$.

### 3.1.3 Leaf node label determination

When an unlabeled sample $x$ is given and the prediction is made on it, $x$ will fall on a leaf node of the tree according to the algorithm. In the tree, the probability that the $x$ is predicted to be class $k$ is

$$\gamma^{(k)}(x) = \frac{1}{N(\mathcal{N}(x))} \sum_{(X,Y) \in \mathcal{N}(x)} \mathcal{I}(Y = k), k = 1, 2, ..., c , \qquad (2)$$

where $\mathcal{I}(\cdot)$ is 1 if $\cdot$ is true and is 0 if $\cdot$ is false; $\mathcal{N}(x)$ is the leaf node where $x$ falls into, $N(\mathcal{N}(x))$ is the sample number of node $\mathcal{N}(x)$. According to the majority voting principle, the prediction of sample $x$ under this tree is

$$\hat{y}(x) = \arg\max_k \{\gamma^{(k)}(x)\} .$$

The prediction of DMRF is the result of majority voting in the base tree, i.e

$$\bar{y}(x) = \arg\max_k \sum_{i=1}^{M} \mathcal{I}(\hat{y}^{(i)}(x) = k) , \qquad (3)$$

where $\hat{y}^{(i)}(x)$ is the predicted value of sample $x$ in the $i$-th decision tree. If there are multiple categories with the same number of votes, we randomly select one of them as the final prediction class.

The pseudo-code of DMRF algorithm and decision tree construction is as follows:

---

**Algorithm1** DMRF classification tree construction process: Tree()

---

1. **Input:** A training set for bootstrap sampling $D_n'$, hyper-parameters $p$, $k_n$, $B_1$, $B_2$.
2. **Output:** A classification tree in DMRF.
3. **While** stopping condition is false **do**

4. Compute the impurity reduction of all possible split points $v_{ij}$ at node $\mathcal{D}$.

5. Select the feature subspace with size $\sqrt{D}$;

6. Do a Bernoulli experiment $B$, $B \sim B(1, p)$;

7. **if** $B = 1$ **then**

8.     use the best split criterion to select the best splitting point.

9. **else**

    The vector $I$ composed of the maximum impurity reduction of each feature in the feature subspace is normalized to $\tilde{I}$, compute the probability $\alpha = soft\max(B_1\tilde{I})$, randomly select a splitting feature according to $M(\alpha)$.

    Assuming this feature is $A_j$.

10.     The impurity reduction vector of feature $A_j$ selected in the previous step is normalized to $\tilde{I}^{(j)}$, compute the probability $\beta = soft\max(B_2\tilde{I}^{(j)})$, randomly select the splitting features according to $M(\beta)$.

11. Split the node $\mathcal{D}$ into left and right child nodes $\mathcal{D}^l$, $\mathcal{D}^r$ by the split features and values obtained.

12. $\mathcal{D}.\text{left} \leftarrow \text{Tree}(\mathcal{D}^l, p, k_n, B_1, B_2)$ and $\mathcal{D}.\text{right} \leftarrow \text{Tree}(\mathcal{D}^r, p, k_n, B_1, B_2)$

13. **end while**

14. **Return:** A classification tree in DMRF.

**Algorithm 2**    DMRF classification algorithm

1. **Input:** Training set $D_n = \{(X_1, Y_1), (X_2, Y_2), ..., (X_n, Y_n)\}$, the number of trees $M \in N^+$, hyper-parameter $p$, $q$, $k_n$, $B_1$, $B_2$, sample $x$.

2. **Output:** DMRF's prediction for sample $x$.

3. **for** $i = 1, 2, ..., M$ **do**

4.     Using bootstrapping with the probability of $q_n$ to get the training set $D_n^{(i)}$.

5.     **if** $D_n^{(i)} = \varnothing$ **then**

6.         Go to line 4 for resampling.

7.      Training a classification tree with the training set $D_n^{(i)}$.

8. **end for**

9. **Return:** Predict the class of $x$ by majority voting.

### 3.1.4 Strong consistency proof of classification DMRF

In this section, we prove the strong consistency of the classification DMRF algorithm, and the detailed proof process is in the appendix.

First, we talk about the consistency definition of classifier. For a classifier sequence $\{g_n\}$, the classifier $g_n$ is obtained by training the data set $D_n = \{(X_1, Y_1), ..., (X_n, Y_n)\}$ which satisfying the distribution $(X, Y)$, and the error rate is

$$L_n = L(g_n) = P(g_n(X, C, D_n) \neq Y \mid D_n),$$

where $C$ is the randomness introduced in the training.

**Definition 3.2**: Given the training set $D_n$ which contain $n$ i.i.d observations, for a certain distribution $(X, Y)$, call classifier $g_n$ is weakly consistent if $g_n$ satisfying

$$\lim_{n \to \infty} EL_n = \lim_{n \to \infty} P(g_n(X, C, D_n) \neq Y) = L^*,$$

where $L^*$ denotes the Bayes risk and $C$ is the randomness introduced in the training. Besides, call $g_n$ is strongly consistent if $g_n$ satisfying

$$P(\lim_{n \to \infty} L_n = L^*) = 1.$$

**Definition 3.3**: A sequence of classifiers $\{g_n\}$ is called weakly(strongly) universally consistent if it is weakly(strongly) consistent for all distributions.

Obviously, the condition of strong consistency is stronger than weak consistency, so strong consistency can derive weak consistency, but vice versa is not necessarily true.

Here are some important lemmas that will be used in the proof.

**Lemma 3.1:** Assume that the classifier sequence $\{g_n\}$ is (universally) strongly consistent, then the majority voting classifier $\bar{g}_n^{(M)}$ (for any value of $M$) is also (universally) strongly consistent.

**Lemma 3.2:** Assume that the classifier sequence $\{g_n\}$ is strongly consistent, the bagging majority voting classifier $\bar{g}_n^M$ (for any value of $M$) is also strongly consistent if $\lim_{n\to\infty} nq = \infty$.

Lemma 3.2 is quoted from Theorem 6 [20]. Refer to [20] for more details.

Lemma 3.1 shows that to prove an ensemble classifier with strong consistency, we only need to prove that its base classifier has strong consistency. The universal strong consistency of ensemble classifiers is obtained from the universal strong consistency of base classifiers. Lemma 3.2 can be regarded as a corollary of Lemma 3.1, which shows that the use of bootstrapping does not affect the consistency of the ensemble algorithm. It is worth noting that Lemma 3.1 alone (without bootstrapping) is sufficient to prove the strong consistency of the DMRF. However, using the whole dataset as the training set to build trees will lead to excessive computational costs and high costs with large sample sizes. Moreover, the similarity among the trees will significantly affect the performance of the algorithm. Therefore, we introduce Lemma 3.2, which adds bootstrapping to reduce the training cost while reducing the similarity between trees.

The strong consistency of a single tree is proved below.

**Lemma 3.3:** Let $g_n$ be a binary tree classifier (that is, a parent node has only two child nodes) obtained by the $n$–sample partitioning rule $\pi_n$, whose each region contains at least $k_n$ points, and $k_n / \log n \to \infty (n \to \infty)$, $A_n(x)$ is the unique cell where the sample $x$ falls into, $\mu(\cdot)$ is the Lebesgue measure of $\cdot$. For all balls $S_r$ with radius $r$ centered at the origin and for all $\gamma > 0$, with probability 1 for all distributions satisfying

$$\lim_{n\to\infty} \mu(\{x : diam(A_n(x) \cap S_r) > \gamma\}) = 0,$$

then $g_n$ corresponding to $\pi_n$ satisfies

$$\lim_{n\to\infty} L(g_n) = L^*$$

with probability one. In other words, $g_n$ is universally strongly consistent.

Lemma 3.3 is quoted from Theorem 21.2 and Theorem 21.8[11], refer to [11] for more details. Lemma 3.3 shows that to prove strong consistency of a tree, we only need to prove that any leaf node is small enough, but the sample size in leaf node is large enough.

Based on the above lemmas, the strong consistency theorem of DMRF algorithm can be obtained:

**Theorem 3.1**: Assume that $X$ is supported on $[0,1]^D$ and has non-zero density almost everywhere, the cumulative distribution function (CDF) of the splitting points is left-continuous at 1 and right-continuous at 0. If $B_1, B_2$ both positive and finite, DMRF is strongly consistent with probability 1 when $k_n/\log n \to \infty$ and $k_n/n \to 0$ as $n \to \infty$.

## 3.2 Regression DMRF

In the last section we discussed the DMRF algorithm for classification, in this section we will discuss the DMRF algorithm for regression.

### 3.2.1 Regression DMRF Algorithm

In classification problems, we choose the Gini index to compute the impurity reduction, while in regression problems, we choose mean squared error(MSE) reduction as the metric for measuring the importance of features and feature values.

Denote the MSE of node $\mathcal{D}$ as

$$MSE(\mathcal{D}) = \frac{1}{N(\mathcal{D})} \sum_{(X,Y) \in \mathcal{D}} (Y - \bar{Y})^2 , \tag{4}$$

where $\bar{Y} = \frac{1}{N(\mathcal{D})} \sum_{(X,Y) \in \mathcal{D}} Y$, i.e., the mean of the samples in this node; $N(\mathcal{D})$ is the sample size of node $\mathcal{D}$. Similar to the classification, when the split point is $v_{ij}$, the MSE reduction is

$$I_{ij} = I(\mathcal{D}, v_{ij}) = MSE(\mathcal{D}) - MSE(\mathcal{D}^l) - MSE(\mathcal{D}^r) , \tag{5}$$

where $\mathcal{D}^l$, $\mathcal{D}^r$ denote the left and right child node of $\mathcal{D}$ splitted by $v_{ij}$.

When making prediction, the predicted value of the tree is the sample mean of the leaf node $A_n(x)$ (where sample $x$ resides), in other words ,

$$\hat{y}(x) = \frac{\sum_{i=1}^{n} Y_i \mathcal{I}(X_i \in A_n(x))}{\sum_{i=1}^{n} \mathcal{I}(X_i \in A_n(x))} = \frac{1}{N(A_n(x))} \sum_{(X,Y) \in A_n(x)} Y , \tag{6}$$

where $N(A_n(x))$ denotes the sample size of $A_n(x)$. The prediction of the forest is the mean of trees, that is

$$\bar{y} = \frac{1}{M} \sum_{i=1}^{M} \hat{y}^{(i)}(x) , \tag{7}$$

where $M$ denotes the tree number of the forest, $\hat{y}^{(i)}(x)$ is the prediction of $i$-th tree towards $x$.

The difference between the regression DMRF and the classification DMRF lies only in the difference in the splitting criteria for the splitting point and the prediction method. To obtain the regression DMRF, we only need to change the impurity reduction criterion to MSE reduction and the majority voting prediction to mean prediction in the classification DMRF.

### 3.2.2 Strong consistency proof of regression DMRF

For a regressor sequence $\{f_n\}$, the regressor $f_n$ is obtained by training the data set $D_n = \{(X_1, Y_1), (X_2, Y_2), ..., (X_n, Y_n)\}$ which satisfying the distribution $(X, Y)$, the MSE of the $f_n$ is

$$R(f_n | D_n) = E[(f_n(X, C, D_n) - f(X))^2 | D_n] , \tag{8}$$

where $C$ is the randomness introduced in the training.

Similar to the classification case, let's first define the strong consistency of a regression problem.

**Definition 3.4**: Given the training set $D_n$ which contain $n$ i.i.d observations, for a certain distribution $(X, Y)$, a sequence of regressors $\{f_n\}$ is called weakly consistent if $f_n$ satisfying

$$\lim_{n \to \infty} E[R(f_n | D_n)] = \lim_{n \to \infty} E[(f_n(X, C, D_n) - f(X))^2] = 0 ,$$

where $f(X) = E[Y | X]$ and $C$ is the randomness introduced in the training. $\{f_n\}$ is called strongly consistent if $f_n$ satisfying

$$\lim_{n\to\infty} R(f_n \mid D_n) = \lim_{n\to\infty} E[(f_n(X,C,D_n) - f(X))^2 \mid D_n] = 0$$

with probability one.

**Definition 3.5:** A sequence of regressors $\{f_n\}$ is called weakly (strongly) universally consistent if it is weakly (strongly) consistent for all distributions of $(X,Y)$ with $EY^2 < \infty$.

**Lemma 3.4:** Assume that the regressor sequence $\{f_n\}$ is (universally) strongly consistent, then the averaged regressor $\overline{f}_n^{(M)}$ (for any value of $M$) is also (universally) strongly consistent.

**Lemma 3.5:** Assume that the regressor sequence $\{f_n\}$ is strongly consistent, the bagging averaged regressor $\overline{f}_n^{(M)}$ (for any value of $M$) is also strongly consistent if $\lim_{n\to\infty} nq = \infty$.

Lemma 3.4 states that if we want to prove a regression ensemble has strong consistency, we only need to prove that its base regressors have strong consistency. Lemma 3.5 is a corollary of Lemma 3.4 and is similar to the classification case. Bootstrapping is not theoretically necessary but is introduced to reduce computational costs and improve algorithm performance. To prove the consistency of the regression, Lemma 3.4 is sufficient.

**Lemma 3.6:** Let $P_n = \{A_{n,1}, A_{n,2}, ...\}$ be a partition of $R^d$ and for each $x \in R^d$ let $A_n(x)$ denote the cell of $P_n$ containing $x$. Assume for any sphere $S$ centered at the origin

$$\lim_{n\to\infty} \max_{A_{n,j}\cap S \neq \emptyset} diam(A_{n,j}) = 0$$

and

$$\lim_{n\to\infty} \frac{|\{j: A_{n,j} \cap S \neq \emptyset\}| \log n}{n} = 0,$$

then the regressor

$$m'_n = \begin{cases} \dfrac{\sum_{i=1}^n Y_i \mathcal{I}(X_i \in A_n(x))}{\sum_{i=1}^n \mathcal{I}(X_i \in A_n(x))}, & \sum_{i=1}^n \mathcal{I}(X_i \in A_n(x)) > \log n \\ 0, & \text{otherwise} \end{cases}$$

is strongly universally consistent.

Lemma 3.6 is quoted from Theorem 23.2 [11], one can refer to [11] for more details.

**Theorem 3.2**：Assume that $X$ is supported on $[0,1]^D$ and has non-zero density almost everywhere, the cumulative distribution function (CDF) of the split points is left-continuous at 1 and right-continuous at 0. If $B_1$, $B_2$ both positive and finite, DMRF is strongly consistent with probability 1 when $k_n/\log n \to \infty$ and $k_n/n \to 0$ as $n \to \infty$.

# 4. Experiment

For the sake of convenience in narration, we use "(SE)" to indicate that the algorithm is from the original paper, like Denil14(SE), BRF(SE) and MRF(SE). We use "(b)" to indicate the algorithm which use the bootstrapping defined in this paper and without separating the structural part and the estimation part, like Denil14(b), BRF(b), MRF(b).

The experiments are divided into three parts: performance test, standard deviation analysis and parameter test. Performance test evaluates the performance of DMRF in classification and regression problems and compares it with three other consistent RF variants (both weakly and strongly consistent), as well as BreimanRF, to demonstrate DMRF's performance. Standard deviation analysis section evaluates the standard deviation of RF variants in classification and regression problems to measure the randomness of different RFs. Parameter test discusses the impact of hyper-parameters $p$, $q$, $B_1$, $B_2$ on the performance of DMRF and provides some recommendations for selecting optimal parameters.

## 4.1 Dataset selection

Our data sets are all from UCI database. Table 1 and Table 2 contain the sample number and feature number of classification and regression data sets respectively, and the classification data set also contains the number of class. In the two tables, we sort the data sets according to the sample number and test the datasets which cover wide range of sample size and feature dimensions in order to show the performance of DMRF. In addition, refer to [14], for missing values of all data sets, we use "-1" padding operation, and no other preprocessing was performed.

Table 1 The description of benchmark classification datasets

| Data sets | Samples | Features | Classes |
|---|---|---|---|
| Blogger | 100 | 6 | 2 |
| Bone marrow | 187 | 39 | 2 |
| Algerian Forest Fires | 244 | 12 | 2 |
| Vertebral | 310 | 6 | 3 |
| Chronic kidney disease | 400 | 25 | 2 |
| Cvr | 435 | 16 | 2 |
| House-votes | 453 | 16 | 2 |
| Wdbc | 569 | 39 | 2 |
| Breast original | 699 | 10 | 2 |
| Balance scale | 625 | 4 | 3 |
| Raisin | 900 | 8 | 2 |
| Vehicle | 946 | 18 | 4 |
| Tic-tac-toe | 958 | 9 | 2 |
| HCV | 1385 | 28 | 4 |
| Winequality(red) | 1599 | 11 | 7 |
| Wireless | 2000 | 7 | 4 |
| Obesity | 2111 | 17 | 7 |
| Ad | 3279 | 1558 | 2 |
| Spambase | 4601 | 57 | 2 |
| Winequality(white) | 4898 | 11 | 7 |
| Page blocks | 5473 | 10 | 5 |
| MFCCs | 7195 | 22 | 4 |
| Mushroom | 8124 | 22 | 7 |
| Ai4i | 10000 | 14 | 3 |
| Letter | 20000 | 16 | 26 |
| Adult | 48842 | 14 | 2 |
| Connect-4 | 67557 | 42 | 3 |

Table 2  The description of benchmark regression datasets

| Datasets | Samples | Features |
|---|---|---|
| ALE | 107 | 6 |
| Alcohol | 125 | 8 |
| Servo | 167 | 4 |
| CSM | 217 | 12 |
| Real estate | 414 | 7 |
| Facebook | 500 | 19 |
| Las Vegas Strip | 504 | 20 |
| Forest fire | 517 | 13 |
| ISTANBUL STOCK | 536 | 8 |
| Qsar fish toxicity | 908 | 7 |
| Qsar BCF Kow | 1056 | 7 |
| Flare | 1389 | 10 |
| Communities | 1994 | 128 |
| Skillcraft | 3395 | 20 |
| Winequality(white) | 4898 | 11 |
| Parkinsons | 5875 | 26 |
| SeoulBikeData | 8760 | 14 |
| Insurance | 9000 | 86 |
| Combined | 9568 | 4 |
| Cbm | 11934 | 16 |

Note: Due to the large value of the CSM, Facebook and SeoulBikeData datasets, the labels are log-transformed.

## 4.2 Baselines

We choose three proposed RF variants with weak consistency, Denil14(SE), BRF(SE) and MRF(SE), as the comparison model of DMRF. Their common feature is that the dataset is divided into the structure part and the estimation part according to the hyper parameter *Ratio* , the structure part is used for split points training and the estimation part for leaf node labels determination.

1) Denil14(SE) randomly selects *m* points of the structure part at each node, then selects the feature subspace with the size of $\min(1+Poisson(\lambda), D)$ without replacing, searches for the optimal splitting point within the range defined by the *m* points preselected (not the entire number of data points).

2) BRF(SE) introduces the first Bernoulli distribution when selecting feature subspace, that means, a feature is randomly selected from the feature set with the probability of $p_1$ as the split feature, or $\sqrt{D}$ features are randomly selected from the feature set with the probability of $1-p_1$ as the candidate feature. The second Bernoulli distribution is introduced in the selection of split values, that means, a value is randomly selected as the split value from the split features with the probability of $p_2$, or the value with the probability of $1-p_2$ is selected from the split features with the largest impurity reduction.

3) MRF(SE) normalize the vector composed of the maximum impurity reduction of each feature when selecting the splitting feature, and convert it into probabilities using softmax function, which is used as multinomial distribution to randomly select the splitting feature. The impurity reduction form a vector corresponding to each value of the obtained splitting feature, normalize and convert this vector into probabilities by softmax function, which is used as multinomial distribution to randomly select the splitting value.

According to our method, we can abandon the separation of the structural part and the estimation part in the three models mentioned above. At the same time, we can add the bootstrapping method defined earlier, they can be Denil14(b), BRF(b) and MRF(b). The experiments will examine the performance of these three models in improving data utilization.

### 4.3 Performance Test Experimental Settings

In the performance experiment, in the above three models, i.e. Denil14(SE), BRF(SE) and MRF(SE), we set $Ratio=0.5$ uniformly. Besides, we set $k_n = 5$ and $M = 100$ according to [14] (The purpose of setting $k_n = 5$ uniformly in this context is to promote extensive tree growth. As long as different algorithms use the same value of $k_n$ for a given data set, it ensures comparability). In Denil14, we set $m = 100$, $\lambda = 10$ according to [14]. Following [13], we set $p_1 = p_2 = 0.05$. As [14] suggested, in MRF, $B_1 = B_2 = 5$ (It should be noted that in this paper, we use $B_1$ and $B_2$ to compute probabilities, while in [14], the authors use $B_1 / 2$ and $B_2 / 2$, they recommend 10 for both $B_1$ and $B_2$, so we set $B_1 = B_2 = 5$). In DMRF, we choose $q = 1 - 1/e$, $p = 0.5$, $B_1 = B_2 = 5$.

### 4.4 Standard Deviation Analysis

The parameters used in the analysis of standard deviation is the same as those used in performance experiment.

### 4.5 Parameter Test Experimental Settings

In the parameter testing experiment, we explore the influence of hyper parameters on DMRF. We focus on $p$, $q$, $B_1$, $B_2$. For $p$ and $q$, the test range we take is [0.05, 0.95] with step size 0.1. For $B_1$, $B_2$ the test range is the integer in [1,10]. In terms of dataset size, we define datasets with less than 500 data points as small, datasets with 500-1000 data points as medium, and datasets with more than 1000 data points as large. For classification problems, accuracy is used as the evaluation metric. For regression problems, negative mean squared error (NMSE) is used as the evaluation metric for ease of observation.

## 5. Results and Discussion

### 5.1 Performance Analysis

In the RF variants, the best performing result in the table is highlighted in bold. To compare the performance of DMRF and BreimanRF, we use "*" to indicate that which is better.

#### 5.1.1 Classification

The evaluation standard of classification problem is accuracy.

Table 3 Accuracy(%) of different RFs on benchmark datasets

| Datasets | DMRF | MRF(SE) | MRF(b) | BRF(SE) | BRF(b) | Denil14(SE) | Denil14(b) | BreimanRF |
|---|---|---|---|---|---|---|---|---|
| Blogger | **81.80*** | 76.20 | 79.90 | 78.30 | 81.20 | 75.80 | 80.50 | 81.40 |
| Bone marrow | **93.96*** | 93.56 | 93.71 | 93.53 | 93.86 | 93.92 | 93.45 | 93.42 |
| Algerian Forest Fires | **93.03*** | 92.25 | 92.16 | 92.71 | 93.71 | 92.45 | 92.46 | 93.00 |
| Vertebral | **84.39** | 83.19 | 83.58 | 83.74 | 84.35 | 82.74 | 82.74 | 84.64* |
| Chronic kidney disease | **98.80*** | 98.13 | 98.60 | 98.23 | 98.65 | 98.23 | 98.48 | 98.77 |
| Cvr | **96.19*** | 95.49 | 95.61 | 95.91 | 96.16 | 95.63 | 95.74 | 95.84 |
| House-votes | **96.17*** | 95.67 | 95.49 | 95.58 | 95.89 | 95.66 | 95.67 | 95.87 |
| Wdbc | **96.25*** | 95.58 | 96.22 | 95.12 | 95.96 | 95.55 | 96.08 | 94.18 |
| Breast original | **95.88** | 95.29 | 95.48 | 95.48 | 95.72 | 94.45 | 94.69 | 96.71* |
| Balance scale | **83.45** | 80.58 | 77.64 | 81.83 | 83.46 | 80.15 | 77.02 | 86.19* |
| Raisin | **86.22*** | 85.47 | 85.94 | 85.60 | 86.02 | 86.16 | 85.53 | 84.98 |
| Vehicle | **75.63*** | 73.24 | 75.60 | 72.79 | 74.44 | 73.04 | 74.55 | 74.46 |
| Tic-tac-toe | 98.27* | 98.47 | **98.85** | 98.18 | 98.08 | 97.82 | 98.38 | 94.37 |
| HCV | **23.84** | 23.55 | 23.66 | 23.25 | 23.28 | 23.28 | 23.22 | 24.88* |
| Winequality(red) | **69.83*** | 62.47 | 69.57 | 62.48 | 69.75 | 62.08 | 69.49 | 64.70 |
| Wireless | **98.36*** | 98.28 | 98.33 | 98.20 | 98.18 | 97.78 | 97.97 | 98.33 |
| Obesity | **78.54** | 24.41 | 77.34 | 71.35 | 78.51 | 73.28 | 77.20 | 94.42* |
| Ad | 97.66* | 96.76 | **97.95** | 94.43 | 97.46 | 94.16 | 96.98 | 97.02 |
| Spambase | **95.18*** | 93.60 | 95.01 | 93.93 | 95.02 | 91.48 | 95.10 | 91.82 |
| Winequality(white) | **69.21*** | 59.93 | 69.20 | 60.65 | 69.44 | 60.07 | 68.71 | 64.02 |
| Page blocks | **97.59*** | 97.44 | 97.56 | 97.17 | 97.45 | 97.28 | 97.44 | 97.06 |
| MFCCs | **98.53*** | 98.02 | 98.50 | 98.02 | 98.47 | 97.83 | 98.36 | 63.57 |
| Mushroom | 57.24* | 59.98 | 47.42 | **62.10** | 58.67 | 58.99 | 48.54 | 47.28 |
| Ai4i | **59.96*** | 59.50 | 57.01 | 59.40 | 59.95 | 59.60 | 56.66 | 56.15 |
| Letter | **89.79** | 89.55 | 89.58 | 83.05 | 89.00 | 81.78 | 87.50 | 96.32* |
| Adult | **86.45*** | 86.28 | 86.13 | 57.57 | 86.44 | 86.19 | 85.57 | 85.98 |
| Connect-4 | 82.18* | 81.96 | **84.08** | 78.86 | 80.77 | 81.28 | 83.44 | 81.46 |

From Table 3, the following conclusions can be drawn:

- In the majority of cases, the (b)-type models show higher accuracy compared to their corresponding (SE)-type models (for example, MRF(b) achieves a 9% higher accuracy than MRF(SE) on the Winequality(white) dataset). This suggests that when the splitting criterion and leaf node label determination process are independent, a significant amount of information may be lost. Determining the leaf node labels based on the samples used to compute the splitting point helps reduce information loss.

- In all datasets, DMRF generally outperforms other RF variations, and the advantage of DMRF over MRF(b) is particularly evident with an improvement of 1% observed on some datasets (such as Obesity and Ai4i).

- In most cases, the accuracy of DMRF is higher than BreimanRF. This can be attributed to the use of the multinomial distribution for randomly sampling splitting values can be seen as a weakened version of optimal splitting, which enhances robustness. It indicates that introducing some randomness in classification tasks can enhance performance.

5.1.2 Regression

The evaluation criterion of regression problem is mean square error.

Table 4 Mean square error(%) of different RFs on benchmark datasets

| Datasets | DMRF | MRF(SE) | MRF(b) | BRF(SE) | BRF(b) | Denil14(SE) | Denil14(b) | BreimanRF |
|---|---|---|---|---|---|---|---|---|
| ALE | **0.0474** | 0.0758 | 0.0494 | 0.0921 | 0.4890 | 0.0931 | 0.0516 | 0.0423* |
| Alcohol | **0.0023*** | 0.0068 | 0.0025 | 0.0049 | 0.0025 | 0.0033 | 0.0024 | 0.0028 |
| Servo | 0.3644 | 0.4248 | **0.2543** | 0.6864 | 0.4247 | 0.7832 | 0.2538 | 0.3114* |
| CSM | **0.0204*** | 0.0711 | 0.0206 | 0.1681 | 0.0209 | 0.0495 | 0.0244 | 0.0239 |
| Real estate | **55.060** | 106.919 | 55.628 | 118.417 | 61.783 | 110.264 | 93.836 | 53.699* |
| Facebook | **0.0510** | 0.7594 | 0.0637 | 1.0288 | 0.0546 | 0.8786 | 0.0546 | 0.0465* |
| Las Vegas Strip | **0.9942** | 1.001 | 1.016 | 1.015 | 1.014 | 1.167 | 1.072 | 0.9826* |
| Forest fire | **2.073*** | 2.159 | 2.191 | 2.166 | 2.164 | 2.166 | 2.156 | 2.232 |

| | DMRF | MRF(SE) | MRF(b) | BRF(SE) | BRF(b) | Denil14(SE) | Denil14(b) | BreimanRF |
|---|---|---|---|---|---|---|---|---|
| ISTANBUL STOCK | **2.14E-04*** | 3.88E-04 | 2.21E-04 | 4.15E-04 | 4.14E-04 | 4.40E-04 | 3.33E-04 | 2.20E-04 |
| Qsar fish toxicity | 0.7794* | 1.047 | **0.7592** | 2.0185 | 0.9218 | 1.9508 | 0.9441 | 1.241 |
| Qsar BCF Kow | **0.7401** | 1.5096 | 0.7695 | 1.4094 | 0.7514 | 1.4111 | 1.1961 | 0.5152* |
| Flare | **0.5219*** | 0.5222 | 0.5937 | 0.5426 | 0.5227 | 0.5684 | 0.6013 | 0.5981 |
| Communities | **0.0203** | 0.0372 | 0.0205 | 0.0521 | 0.2049 | 0.0499 | 0.0244 | 0.0182* |
| Skillcraft | **4.37E-08** | 6.54E-08 | 4.39E-08 | 6.95E-08 | 5.34E-08 | 7.02E-08 | 5.64E-08 | 4.23E-08* |
| Winequality(white) | **0.3562*** | 0.6177 | 0.3601 | 0.7749 | 0.3815 | 0.7311 | 0.3949 | 0.3640 |
| Parkinsons | **0.0013*** | 0.0029 | 0.0014 | 0.0078 | 0.0016 | 0.0045 | 0.0016 | 0.0013* |
| SeoulBikeData | **0.2059** | 0.6722 | 0.207 | 1.4543 | 0.3664 | 1.2996 | 0.8322 | 0.1782* |
| Insurance | 0.0585 | 0.0559 | 0.0596 | 0.0561 | **0.0549** | 0.0561 | 0.0601 | 0.0554* |
| Combined | **11.700** | 15.371 | 11.716 | 48.995 | 12.445 | 26.054 | 14.445 | 10.819* |
| Cbm | **1.00E-06** | 8.00E-06 | **1.00E-06** | 5.55E-05 | 1.50E-05 | 5.61E-05 | 2.00E-06 | 7.95E-07* |

From Table 4, the following conclusions can be drawn:

- Similar to the classification case, in the majority of cases, the (b)-type models outperform their corresponding (SE)-type models (for example, there is an 48% reduction in MSE on the "Real estate" dataset). This suggests that determining the leaf node labels based on the samples used to compute the splitting point can better utilize information compared to the independent processes.
- In all datasets, DMRF generally shows the best performance among RF variations, but the advantage of DMRF over MRF(b) is not significant ,
- In most cases, the MSE of DMRF is larger than that of BreimanRF. This is because MSE amplifies the impact of noise, and the randomness introduced by the multinomial distribution makes DMRF perform worse in regression compared to BreimanRF. In contrast, DMRF is better suited for classification tasks.

### 5.2 Standard Deviation Analysis

Given the inherent randomness in the models used for the experiments, it is essential to compare their levels of randomness. In this case, we will evaluate the randomness using the standard deviation as a metric.

Table 5 Standard deviation of different RFs on classification and regression datasets

| Task | Datasets | DMRF | MRF(SE) | MRF(b) | BRF(SE) | BRF(b) | Denil14(SE) | Denil14(b) | BreimanRF |
|---|---|---|---|---|---|---|---|---|---|
| | Blogger | 0.9189 | 1.4757 | 2.0789 | 2.3593 | 2.0439 | 1.3984 | 2.0138 | 2.1832 |
| | Algerian Forest Fires | 0.2907 | 0.4842 | 0.7431 | 0.5495 | 0.4650 | 0.3465 | 0.8440 | 0.5922 |
| | Wdbc | 0.2153 | 0.3031 | 0.3751 | 0.4283 | 0.3108 | 0.1860 | 0.2476 | 0.4878 |
| Classif-ication | Tic-tac-toe | 0.1494 | 0.1980 | 0.2642 | 0.1492 | 0.3619 | 0.1690 | 0.2748 | 0.2433 |
| | Winequality(Red) | 0.3718 | 0.4873 | 0.6304 | 0.5572 | 0.4748 | 0.4111 | 0.5347 | 0.4871 |
| | Wireless | 0.1054 | 0.0919 | 0.1061 | 0.0422 | 0.1752 | 0.0577 | 0.1414 | 0.1006 |
| | Winequality(white) | 0.3107 | 0.2374 | 0.2595 | 0.3072 | 0.2987 | 0.1558 | 0.2839 | 0.2856 |
|  | Connect-4 | 0.0629 | 0.0349 | 0.0627 | 0.0594 | 0.0591 | 0.0262 | 0.0745 | 0.0556 |
|  | ALE | 0.002513 | 0.002303 | 0.002802 | 0.002485 | 0.002053 | 0.001867 | 0.003254 | 0.002637 |
|  | Servo | 0.0477 | 0.0158 | 0.0200 | 0.0568 | 0.0846 | 0.0201 | 0.0790 | 0.0601 |
|  | Las Vegas Strip | 0.0094 | 0.0038 | 0.0162 | 0.0021 | 0.3846 | 0.0011 | 0.0333 | 0.0173 |
| Regre- | Qsar fish toxicity | 0.0105 | 0.0117 | 0.0081 | 0.0116 | 0.0205 | 0.0087 | 0.0131 | 0.0247 |
| ssion | Flare | 0.003023 | 0.004029 | 0.011366 | 0.001570 | 0.004659 | 0.000626 | 0.0101 | 0.005616 |
|  | Winequality(white) | 0.003234 | 0.002213 | 0.003777 | 0.001211 | 0.002729 | 0.001602 | 0.005282 | 0.003536 |
|  | Insurance | 1.91E-04 | 1.64E-04 | 2.27E-04 | 9.17E-06 | 1.18E-04 | 7.57E-06 | 4.22E-04 | 1.85E-04 |
|  | Combined | 0.0738 | 0.0353 | 0.0595 | 0.4213 | 0.0636 | 0.1011 | 0.0709 | 0.0689 |

Table5 shows the standard deviations of 10-fold cross-validation results computed 10 times for 8 classification datasets and 8 regression datasets. It can be observed that, in both classification and regression task, in most cases, the (b)-type models show larger standard deviations compared to their corresponding (SE)-type models (for example, MRF(b) has a larger standard deviation than MRF(SE) on the Blogger and ALE datasets). This indicates that, under similar conditions, using bootstrapping to get training sets introduces greater randomness compared to divide the dataset into structure part and estimation part.

Furthermore, whether in classification or regression, in cases with small and medium sample sizes, DMRF tends to be more stable than MRF(b) and BreimanRF (for example, on the Wdbc and Qsar fish toxicity datasets). However, with large sample sizes, DMRF exhibits higher randomness compared to MRF(b) and BreimanRF (for example, on the Winequality(white), Connect-4, Insurance, and Combined datasets). This can be attributed to the introduction of Bernoulli and multinomial distributions in the process of finding splitting points in DMRF, which results in higher levels of randomness compared to MRF(b) and BreimanRF. Under small sample size cases, adding appropriate randomness helps increase the robustness of the model. However, in large sample size cases, the difference in randomness is amplified, leading to higher standard deviations for DMRF compared to MRF(b) and BreimanRF.

### 5.3 Parameter Analysis

In this section, we explore the influence of hyper parameters on DMRF.

### 5.3.1 The effect of $p$, $q$

We investigate $p$, $q$ under $B_1 = B_2 = 5$ as [14] recommended.

1）Classification

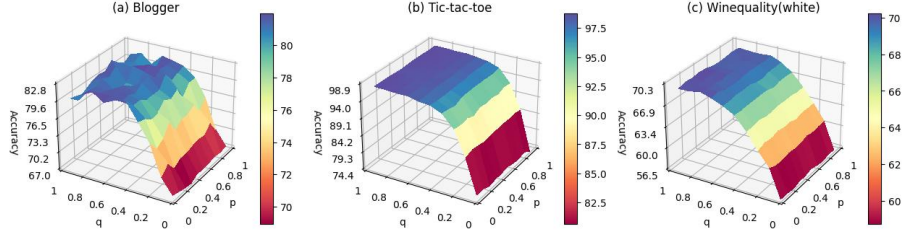

Fig. 1　Accuracy(%) of the DMRF under different p, q values

Fig. 1 shows the performance of DMRF on three classification datasets with small, medium, and large sample sizes under $B_1 = B_2 = 5$. It can be observed that for the same $p$, the accuracy of three datasets at $q = 0.63$ has two situations: close to maximum and stable, or decreasing. Through analysis, it is known that when $q$ is too small, the sampling probability of each sample is too low, resulting in insufficient training of trees; as $q$ increases, the number of sampled samples increases, and the training of trees gradually becomes sufficient, resulting in the performance of DMRF increasing; However, when the number of samples reaches a certain value, the performance improvement slows down, and there will be a situation where the accuracy tends to be stable or even starts to decrease. The reason for the decrease is that the number of samples taken exceeds an appropriate value, resulting in high similarity between the training sets of trees, which affects the overall performance. Since the optimal value of $q$ is around 0.63, we set $q = 1 - 1/e (\approx 0.6322)$ to reduce computational burden while obtaining the optimal parameter.

It is worth noting that the bootstrapping used in this paper can be seen as a general form of the non-repeated resampling standard bootstrapping in the case of large samples. In fact, if we take $n$ non-repeated samples of an $n$-sample dataset with equal probability, the probability of each sample being selected is $1 - (1 - 1/n)^n \to 1 - 1/e (n \to \infty)$. This is also one of the reasons why we chose the $q = 1 - 1/e$.

Under the same $q$, it can be observed that the accuracy increases initially with an increase in $p$ and then decreases, reaching an optimal value around $p = 0.5$. In DMRF, if $p=0$, the selection of splitting points at each node depends on the random selection of two multinomial distributions. If $p=1$, DMRF selects the optimal split point from the feature subspace, which is similar to BreimanRF. It can be seen that introducing some randomness when selecting split points in the feature subspace can enhance performance. Additionally, we can conclude that the algorithm is more sensitive to the parameter $q$ compared to $p$, indicating that the number of training samples is more important than the method of split point selection.

2）regression：

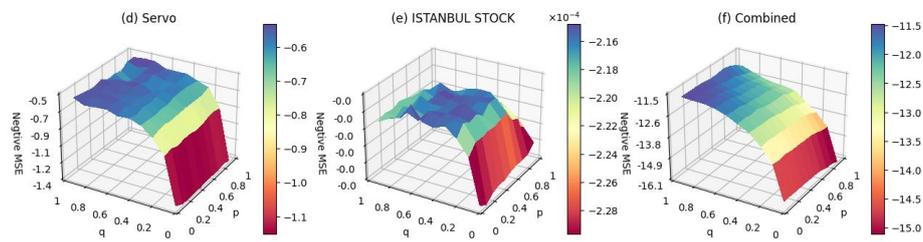

Fig. 2 Negative mean square error of the DMRF under different p, q values

Fig. 2 shows the performance of DMRF on three regression datasets with small, medium, and large sample sizes under $B_1 = B_2 = 5$. It can be observed that the regression situation is similar to the classification: under the same $p$, the NMSE increases with the increase of $q$ and reaches its maximum at around 0.63 before stabilizing or beginning to decrease. Under the same $q$, the NMSE initially increases with the increase of $p$, then decreases or stabilizes after reaching a certain point. Therefore, the optimal $q$ is considered to be $1-1/e$ and the optimal $p$ value is considered to be 0.5. Additionally, it's still observed that the DMRF algorithm is more sensitive to parameter $q$ than to parameter $p$.

5.3.2 The effect of $B_1$, $B_2$

1）classification

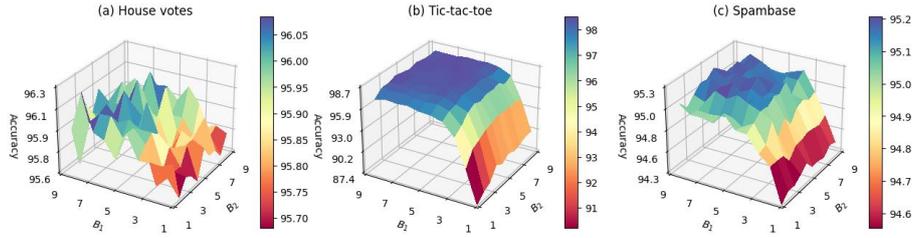

Fig. 3 Accuracy(%) of the DMRF under different B1, B2 values

Fig. 3 shows the impact of $B_1$ and $B_2$ on the DMRF algorithm under $q = 1-1/e$ and $p = 0.5$ for three classification datasets with small, medium, and large sample sizes. Under the same $B_2$, the accuracy increases slightly as $B_1$ increases from 1, reaching a stable or top point at around $B_1 = 5$. Under the same $B_1$, the accuracy starts to increase as $B_2$ increases from 1 and stabilizes or be the top at around $B_2 = 5$. The reason is that when $B_1$ is close to 1, the probabilities of each feature are not significantly different from each other, making it difficult to sample the optimal features. As $B_1$ grows, the differences in probabilities between features become larger, and more important features tend to be selected, improving the DMRF's performance. However, when $B_1$ grows to a certain extent, the selection of features becomes similar to selecting the optimal feature, leading to the similarity between base decision trees being too high and causing a decrease in performance. The situation for $B_2$ is similar to that of $B_1$.

It can also be seen from the figure that the DMRF is not sensitive to $B_2$ but is more sensitive to $B_1$. Since $B_1$ affects the selection of splitting features and $B_2$ affects the selection of split values, this can indicate that the impact of the splitting feature is greater. This also makes sense because the splitting value is obtained based on the splitting feature, so in general, the influence of the splitting feature is larger compared to the splitting value.

2）regression

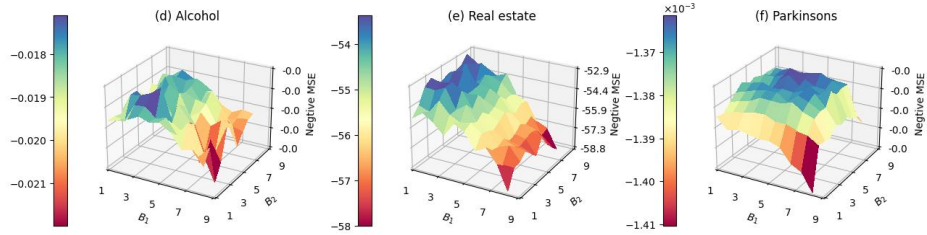

Fig. 4 Negative mean square error of the DMRF under different B1, B2 values

Fig. 4 shows the impact of $B_1$ and $B_2$ on the DMRF algorithm under $q = 1 - 1/e$ and $p = 0.5$ for three regression datasets with small, medium, and large sample sizes. Unlike the classification case, under the same $B_2$, the NMSE increases as $B_1$ decreases from 10 in general. After reaching around $B_1 = 5$, NMSE stabilizes or starts to decrease. Under the same $B_1$, NMSE starts to increase as $B_2$ increases from 1 and stabilizes or slightly decreases at around $B_2 = 5$. The reason is that when $B_1$ is too large, the probability of selecting the optimal feature is much higher than that of other features, resulting in high similarity between trees and poor performance. As $B_1$ decreases, more randomness is introduced, preserving the high performance of trees while increasing diversity, improving the algorithm's performance. When $B_1$ is close to 1, the probabilities of each feature are not significantly different from each other, making it difficult to sample optimal features, leading to the poor performance of trees and the algorithm.

The situation for $B_2$ is similar to that of $B_1$. Futhermore, it can be observed from the Fig 4. that DMRF is more sensitive to $B_1$ than $B_2$, which is similar to that of the classification. Since $B_1$ affects the selection of splitting features and $B_2$ affects the selection of split values, it can be concluded that the impact of the splitting feature is greater.

5.3.3 The effect of number of trees

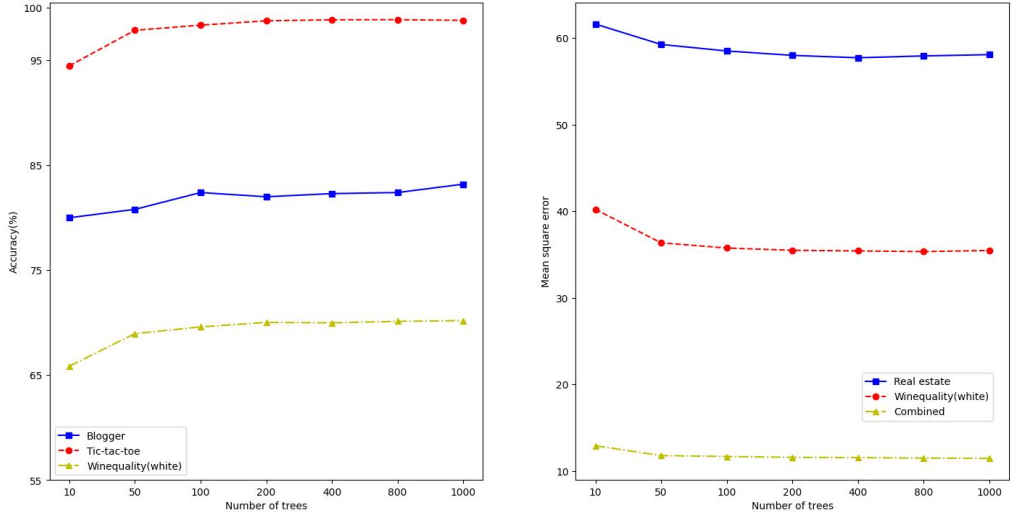

Fig. 5 Performance of DMRF under different number of trees

The left side of Fig 5. shows the performance trends of the Blogger, Tic-tac-toe, and Winequality (white) datasets in terms of accuracy as the number of trees (i.e. $M$) increases in classification. The right side shows the performance trends of the Real estate, Winequality (white), and Combined datasets in terms of mean squared error (MSE) as $M$ increases in regression. It is worth noting that the MSE of the Winequality (white) dataset has been multiplied by 100 for ease of observation.

In the classification case, it can be observed that as $M$ increases, DMRF shows an upward trend in accuracy for all three datasets and gradually converges after reaching 100 trees. Similarly, in the regression case, as $M$ increases, DMRF shows a downward trend in MSE for all three datasets and gradually converges after reaching 100 trees.

This demonstrates that DMRF, as a method of Bagging, improves its performance gradually with an increasing number of base learners until convergence.

5.3.4 Cross-validation

In this section, we use cross validation to determine the optimal parameters for DMRF, MRF(b), BRF(b), and Denil14(b) on various classification and regression datasets. For all models, set $q \in \{0.2, 0.4, 1-1/e, 0.8\}$. For DMRF and MRF(b), set $B_1$, $B_2 \in \{2, 5, 8, 10\}$, $p \in \{0.1, 0.3, 0.5, 0.8\}$. For BRF(b), set $p_1, p_2 \in \{0.01, 0.05, 0.35, 0.65\}$. For Denil14(b), set $\lambda \in \{1, 5, 10, 15\}$, $m \in \{50, 100, 200, 500\}$.

Table 6 Results of cross-validation of different RFs

| Task | Dataset | Algorithm | Performance | Optimal parameters |
|---|---|---|---|---|
| Classification | Blogger | DMRF | **84.8** | $B_1=5, B_2=8, p=0.3, q=0.8$ |
| | | MRF(b) | 84 | $B_1=8, B_2=5, q=0.8$ |
| | | BRF(b) | 84.2 | $p_1=0.05, p_2=0.35, q=0.8$ |
| | | Denil14(b) | 83.2 | $\lambda=10, m=500, q=0.8$ |
| | Vertebral | DMRF | **84.76** | $B_1=2, B_2=2, p=0.5, q=0.8$ |
| | | MRF(b) | 84.58 | $B_1=2, B_2=10, q=1-1/e$ |
| | | BRF(b) | 84.6 | $p_1=0.05, p_2=0.35, q=1-1/e$ |
| | | Denil14(b) | 84.45 | $\lambda=10, m=200, q=0.4$ |
| | House votes | DMRF | **96.26** | $B_1=2, B_2=8, p=0.3, q=0.8$ |
| | | MRF(b) | 96.16 | $B_1=2, B_2=2, q=0.4$ |
| | | BRF(b) | 96.14 | $p_1=0.05, p_2=0.65, q=1-1/e$ |
| | | Denil14(b) | 95.85 | $\lambda=10, m=50, q=0.4$ |
| Regression | ALE | DMRF | **0.0436** | $B_1=8, B_2=10, p=0.5, q=0.4$ |
| | | MRF(b) | 0.0443 | $B_1=8, B_2=2, q=1-1/e$ |
| | | BRF(b) | 0.0451 | $p_1=0.01, p_2=0.01, q=0.4$ |
| | | Denil14(b) | 0.0443 | $\lambda=5, m=50, q=0.4$ |
| | Real estate | DMRF | **51.54** | $B_1=5, B_2=2, p=0.3, q=0.8$ |
| | | MRF(b) | 52.83 | $B_1=5, B_2=2, q=0.8$ |
| | | BRF(b) | 52.49 | $p_1=0.01, p_2=0.65, q=0.8$ |
| | | Denil14(b) | 58.56 | $\lambda=1, m=50, q=0.8$ |

| | | | |
|---|---|---|---|
| | DMRF | **0.5099** | $B_1 = 5, B_2 = 5, p = 0.5, q = 0.2$ |
| Flare | MRF(b) | 0.5227 | $B_1 = 2, B_2 = 2, q = 0.2$ |
| | BRF(b) | 0.5102 | $p_1 = 0.01, p_2 = 0.01, q_n = 0.2$ |
| | Denil14(b) | 0.5156 | $\lambda = 1, m = 100, q = 0.2$ |

In both classification and regression, comparing the performance of DMRF with MRF(b), BRF(b), and Denil14(b) under the optimal parameters, it can be seen from the Table 6 that DMRF outperforms the other models. Compared to the MRF(b), DMRF shows improvements in performance. The reason behind this can be analyzed as follows: Although the MRF(b) uses the softmax function to convert features and feature importance into probabilities and samples splitting points using a multinomial distribution, it considers the entire feature space. The higher the importance of a feature or feature value, the higher its probability. Additionally, the softmax function amplifies the differences between features or feature values. Therefore, the probability of sampling the optimal feature and optimal feature value remains the highest. This can be considered as a weakened version of optimal splits in the full feature space. Although this approach improves performance, it reduces the diversity among trees.

On the other hand, DMRF selects optimal splits in the feature subspace based on probabilities and performs multinomial distribution sampling, which increases the diversity among trees and thus improves performance.

### 5.4 Computational complexity analysis

Assume that the data set has $n$ samples and $D$ features, we prepare to build $M$ trees. The complexity of random sampling is not considered below.

The best case for tree construction is complete balanced growth, in this case, the depth of the tree is $\mathcal{O}(\log n)$. All samples of each layer of DMRF are involved in the calculation, and the number of features calculated is $\sqrt{D}$. Therefore, the complexity of building a DMRF tree is $\mathcal{O}(\sqrt{D}n \log n)$, so the complexity of DMRF is $\mathcal{O}(\sqrt{D}nM \log n)$.

In the same way, the complexity of BreimanRF is $\mathcal{O}(\sqrt{D}nM \log n)$. The feature subspace size of Denil14 is $\min(1 + Poisson(\lambda), D)$, and the optimal split point is searched in the pre-selected samples of $m(m<n)$ samples, so its complexity is $\mathcal{O}(\min(1+Poisson(\lambda),D)\cdot mM\log n)$. BRF introduces two Bernoulli distributions when choosing split points, the average number of features calculated at each layer is $p_1+(1-p_1)\sqrt{D}$, and the average number of samples calculated at each layer is $(1-p_2)n$, so the complexity is $\mathcal{O}((p_1+(1-p_1)\sqrt{D})(1-p_2)nM\log n)$. MRF introduces two multinomial distributions when selecting split points, and all features and samples at each node are involved in the calculation, so the complexity is $\mathcal{O}(DnM\log n)$.

Table 7 Computational complexity of RFs

| RF variants | Complexity |
|---|---|
| BreimanRF | $\mathcal{O}(\sqrt{D}nM\log n)$ |
| BRF | $\mathcal{O}((p_1+(1-p_1)\sqrt{D})(1-p_2)nM\log n)$ |
| DMRF | $\mathcal{O}(\sqrt{D}nM\log n)$ |
| Denil14 | $\mathcal{O}(\min(1+Poisson(\lambda),D)\cdot mM\log n)$ |
| MRF | $\mathcal{O}(DnM\log n)$ |

From Table5, it can be seen that MRF(b) has the highest complexity, followed by DMRF and BreimanRF. Due to the sampling process involved in selecting split points, DMRF will generally take slightly longer than BreimanRF in most cases. Due to the generally small values of $p_1$, $p_2$, in most cases, the complexity of BRF is slightly lower than that of DMRF and BreimanRF. As for Denil14(b), the complexity ranking is determined by the values of $\lambda$, $m$.

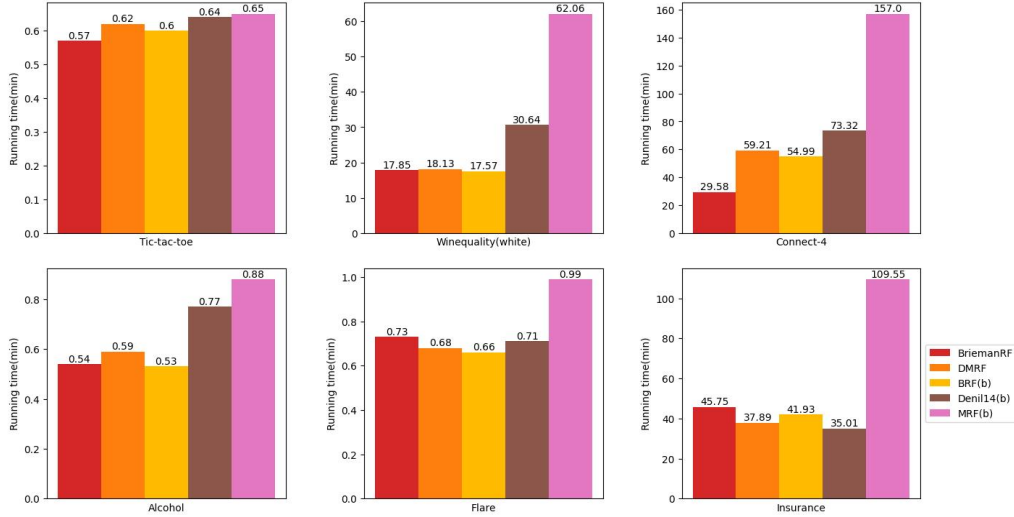

Fig. 6 Running time in one iteration of cross-validation for different models

Fig. 6 shows the running time of one iteration of cross-validation for three classification datasets (Tic-tac-toe, Winequality(white), Connect-4) and three regression datasets (Alcohol, Flare, Insurance) under parameters mentioned in Section 4.3. It can be observed that BRF(b) has shorter runtime in both classification and regression tasks, while MRF(b) has longer runtime in both tasks. As we analyzed earlier, in most cases, BreimanRF has shorter runtime compared to DMRF.

It is worth noting that in Connect-4, a big dataset for classification, Denil14(b) has longer runtime compared to DMRF, BRF(b), and BreimanRF, while in Insurance, a big dataset for regression, Denil14(b) has shorter runtime compared to DMRF, BRF(b), and BreimanRF. This is because Connect-4 has 42 features, and the average size of Denil14(b)'s feature subspace is 11, while the average size of feature subspaces for DMRF, BRF(b), and BreimanRF is 6. Although Denil14(b) only selects 100 points to compute split points, the experimental results indicate that the impact of feature subspace size on runtime in Connect-4 is greater than calculating splitting points for 100 samples each time.

In Insurance, the number of features is 86, and the average size of Denil14(b)'s feature subspace is 11, while the average size of feature subspaces for DMRF, BRF(b), and BreimanRF is 9, which is not significantly different. However, because Denil14(b) only selects 100 points to compute splitting points, it has a faster speed.

## 6. Conclusions and future works

The main contributions of this paper are as follows:

①By modifying the condition for the number of samples in leaf nodes, the weak consistency proofs in previous RF variants have been improved to strong consistency proofs. The previously proposed weak consistency models, such as Denil14(SE), BRF(SE), and MRF(SE), have been enhanced to models with strong consistency in probability, namely Denil14(b), BRF(b), and MRF(b).

②We introduces a novel algorithm called DMRF, which combines Bernoulli and multinomial distributions. DMRF utilizes a modified bootstrapping method to obtain training sets for base trees and uses the combination of Bernoulli and multinomial distributions to determine the splitting points during tree construction. This approach increases diversity while maintaining high-performance base trees.

③The experiments indicate that DMRF outperforms MRF(b) and BreimanRF in classification tasks. However, in regression tasks, DMRF performs better than MRF(b) but the difference is not significant. In most cases, DMRF's performance is not as good as BreimanRF, suggesting that DMRF is more suitable for classification tasks.

④ In terms of standard deviation, DMRF has lower standard deviation than MRF(b) and BreimanRF on small and medium-sized datasets. However, on large datasets, DMRF is more likely to have a higher standard deviation compared to MRF(b) and BreimanRF. In terms of time complexity, DMRF has the same complexity as BreimanRF. The complexity of BRF(b) and Denil14(b) is determined by the parameter settings, while MRF(b) has the highest complexity.

The main advantages of DMRF lie in its strong theoretical properties, excellent performance, and lower complexity(same as BreimanRF). It shows clear advantages on small sample datasets. However, one limitation is that DMRF shows higher randomness than BreimanRF on large dataset. Future research can focus on addressing this limitation of increased randomness in DMRF for large sample cases.

**Acknowledgements**

We would like to thank the Faculty of Science and the Faculty of Information Technology at Beijing University of Technology for their support of this paper. We also appreciate the valuable comments and suggestions from the editors and reviewers

**Author contributions**

JHC: Investigation, Methodology, Coding, Writing - review & editing; FL: Supervision, Methodology, Funding acquisition, review & editing; XLW: Supervision, Methodology, Validation, Writing - review & editing; All authors read and approved the final manuscript.

**Funding**

Not applicable.

**Availability of data and materials**

The experimental data are available at https://archive.ics.uci.edu/ml/index.php.

# Declaration

### Ethics approval and consent to participate

Not applicable.

### Consent for publication

Not applicable.

### Competing interests

The authors declare that they have no competing interests.

# Appendix

### The proof of Lemma 3.1:

Denote $g^*(x)$ as the Bayes classifier, then the Bayes risk is

$$L^* = P(g^*(x) \neq Y).$$

Denote

$$A = \{k \mid \gamma^{(k)}(x) = \max\{\gamma^{(k)}(x)\}\},$$

$$B = \{k \mid \gamma^{(k)}(x) < \max\{\gamma^{(k)}(x)\}\}.$$

Then

$$P(\bar{g}_n^{(M)}(x, C, D_n) \neq Y \mid D_n)$$

$$= \sum_k P(\bar{g}_n^{(M)}(x, C, D_n) = k \mid D_n) P(Y \neq k \mid D_n)$$

$$\leq L^* \cdot \sum_{k \in A} P(\bar{g}_n^{(M)}(x, C, D_n) = k \mid D_n) + \sum_{k \in B} P(\bar{g}_n^{(M)}(x, C, D_n) = k \mid D_n)$$

so it is sufficient to prove that the limit of the latter term is 0 for all $k \in B$.

For $\forall k \in B$,

$$P(\bar{g}_n^{(M)}(x, C, D_n) = c \mid D_n)$$

$$= P(\sum_{i=1}^M \mathcal{I}(g_n(x, C^{(i)}, D_n) = k) > \max_{l \neq k}\{\sum_{i=1}^M \mathcal{I}(g_n(x, C^{(i)}, D_n) = l)\} \mid D_n)$$

$$\leq P(\sum_{i=1}^M \mathcal{I}(g_n(x, C^{(i)}, D_n) = k) \geq 1 \mid D_n)$$

$$\leq E(\sum_{i=1}^{M} \mathcal{I}(g_n(x,C^{(i)},D_n)=k)\,|\,D_n)$$

$$= MP(g_n(x,C,D_n)=k)\,|\,D_n) \to 0\,(n\to\infty).$$

$\square$

**The proof of Lemma 3.4:**

Every base tree is strongly consistent, i.e.,

$$\lim_{n\to\infty} R(f_n\,|\,D_n) = \lim_{n\to\infty} E[(f_n(X,C^{(i)},D_n)-f(X))^2\,|\,D_n] = 0\,,\, i\in\{1,2,...,M\}.$$

Then

$$R(\overline{f}_n^{(M)}\,|\,D_n)$$

$$= E[(\frac{1}{M}\sum_{i=1}^{M}f_n(X,C^{(i)},D_n)-f(X))^2\,|\,D_n]$$

$$\overset{(c)}{\leq} \frac{1}{M}\sum_{i=1}^{M}E[(f_n(X,C^{(i)},D_n)-f(X))^2\,|\,D_n] \to 0\,(n\to\infty).$$

where $C^{(i)}$ is the randomness introduced in $i$-th tree building, (c) uses Cauchy inequality.

$\square$

**The proof of Theorem 3.1:**

First, we proof the number of samples in each leaf node of DMRF tree has at least $k_n$ with probability 1 when $n\to\infty$.

Due to the randomness of the split point selection, the final selected split point can be regarded as a random variable $W$, which follows the uniform distribution on [0,1], and its cumulative distribution function is

$$F_W(x) = x, x\in[0,1].$$

For $\forall m\in N^+, \varepsilon>0$ and a certain $0<\eta<1$, the smallest child node after the root node splits according to a splitting feature is denoted as $M_1 = \min(W,1-W)$, then we have

$$P(M_1 \geq \eta^{1/m}) = P(\eta^{1/m} \leq W \leq 1-\eta^{1/m})$$

$$= F_W(1-\eta^{1/m}) - F_W(\eta^{1/m})$$

$$= 1 - \eta^{1/m} - \eta^{1/m}$$

$$= 1 - 2\eta^{1/m}.$$

Without loss of generality, we can normalize the value of all attributes to range $[0,1]$ for each node. If the feature is continuously split $m$ times (i.e. the tree grows to the $m$-th layer), the probability that the smallest child node in $m$-th layer has the size at least $\eta$ is

$$P(M_m \geq \eta) = (1 - 2\eta^{1/m})^m.$$

In this case, if $0 < \eta < \{\frac{1}{2}[1-(1-\varepsilon)^{\frac{1}{K}}]\}^K$, then

$$P(M_m \geq \eta) = (1 - 2\eta^{1/m})^m > 1 - \varepsilon.$$

The above results are based on the fact that the same feature is selected for each split. In fact, if different features are split at different layers, $P(M_m \geq \eta)$ will be greater, we still have

$$P(M_m \geq \eta) > 1 - \varepsilon.$$

This indicates that the size of each node is $\eta$ in $m$-th layer with the probability at least $1 - \varepsilon$.

Since $X$ has a non-zero density function, each node in the $m$-th layer of the tree has a positive metric with respect to $\mu_X$. Define

$$\zeta = \min_{\mathcal{N}: a\ leaf\ at\ m-th\ level} \mu_X(\mathcal{N}),$$

$\zeta > 0$ because the measure of each leaf node is positive and the number of leaf nodes is finite.

The number of samples in the training set is $n$, and the number of samples in the leaf node $\mathcal{N}$ is $B(n, \zeta)$, then

$$P(N(\mathcal{N}) < k_n) = P(N(\mathcal{N}) - n\zeta < k_n - n\zeta)$$

$$\overset{(a)}{=} P(|N(\mathcal{N}) - n\zeta| > |k_n - n\zeta|)$$

$$\overset{(b)}{\leq} \frac{n\zeta(1-\zeta)}{|k_n - n\zeta|^2}$$

$$= \frac{\zeta(1-\zeta)}{n|\frac{k_n}{n} - \zeta|^2} \to 0(n \to \infty).$$

(a) is based on the fact that $k_n/n \to 0$ as $n \to \infty$, so $k_n - n\zeta < 0$ if $n \to \infty$. (b) uses Chebyshev's inequality. This suggests that the probability of reaching the stop condition will converge to 0 as $n \to \infty$, which means that can split infinitely many times with probability 1.

It is sufficient to show that it satisfies the conditions of Lemma 3.3 with probability 1. Obviously, we just need to prove that $diam(A_n(x)) \to 0$ as $n \to \infty$ with probability 1. Let $V(i)$ denote the size of the $i$-th feature of $A_n(x)$, we only need to show that $E[V(i)] \to 0$ for all $i \in \{1, 2, ..., D\}$.

Without loss of generality, at each node, we will scale each feature to [0, 1].

First, we define the following events: $E_1 = \{\text{i-th feature is a candidate feature}\}$, $E_2 = \{\text{use optimal split criterion to get splitting point}\}$, $E_3 = \{\text{i-th feature is a splitting feature}\}$. For a given $i$, denote the largest size among its child nodes as $V^*(i)$.

Let $W_i$ be the position of the splitting point, then $W_i | E_3 \sim U[0,1]$ and

$$V^*(i) | E_3 = \max(W_i | E_3, 1 - (W_i | E_3)) \sim U[\frac{1}{2}, 1],$$

so we have

$$E[V^*(i) | E_3] = \frac{3}{4}.$$

When $\bar{E}_2$ happens, during the process of selecting the splitting feature, the normalized vector of impurity reduction (denoted as $\hat{I}$, which is an $n$-dimensional vector) is considered. When the i-th element of $\hat{I}$ is 0 and all other elements are 1, the probability of selecting the i-th feature as the splitting feature is minimized. Therefore,

$$P(E_3 | \bar{E}_2) \geq \frac{1}{1+(\sqrt{D}-1)e^{B_1}} \overset{\Delta}{=} p_1,$$

and
$$P(E_3) = P(E_2)P(E_3 | E_2) + P(\bar{E}_2)P(E_3 | \bar{E}_2) \geq P(\bar{E}_2)P(E_3 | \bar{E}_2) \geq (1-p)p_1.$$

So,
$$E[V(i) | E_1] \leq P(E_3 | E_1) \cdot E[V(i) | E_1, E_3] + P(\bar{E}_3 | E_1)E[V(i) | E_1, \bar{E}_3]$$

$$= P(E_3)E[V(i) | E_3] + (1 - P(E_3)) \cdot 1$$

$$\leq P(E_3)E[V^*(i) | E_3] + 1 - P(E_3)$$

$$= P(E_3) \cdot \frac{3}{4} + 1 - P(E_3)$$

$$= 1 - \frac{1}{4}P(E_3)$$

$$\leq 1 - \frac{(1-p)p_1}{4}.$$

Thus, it can be inferred that
$$E[V(i)] \leq P(E_1) \cdot E[V(i) | E_1] + P(\bar{E}_1)E[V(i) | \bar{E}_1]$$

$$= \frac{\sqrt{D}}{D} \cdot E[V(i) | E_1] + (1 - \frac{\sqrt{D}}{D}) \cdot 1$$

$$\leq \frac{1}{\sqrt{D}}(1 - \frac{(1-p)p_1}{4}) + 1 - \frac{1}{\sqrt{D}}$$

$$= 1 - \frac{(1-p)p_1}{4\sqrt{D}}$$

$$\stackrel{\Delta}{=} A \quad (\text{denote } A = 1 - \frac{(1-p)p_1}{4\sqrt{D}}).$$

The above process is the result of one time split. If the $i$-th feature is splited $m$ times, the following formula can be obtained by iterating the above formula continuously:

$$E[V(i)] \leq A^m.$$

We have proven that $m \to \infty (n \to \infty)$ with probability 1, so the strong consistency of DMRF can be obtained with probability 1.

In summary, $diam(A_n(x)) \to 0 (n \to \infty)$ with probability 1, DMRF tree is strongly consistent with probability 1. By lemma 3.2, DMRF algorithm has strong consistency with probability 1.

□

**The proof of Theorem 3.2:**

By lemma 3.4, the strong consistency of DMRF in regression problem is based on the strong consistency of trees. The following proves the strong consistency of the base regression tree.

By lemma 3.6, if we prove

$$\lim_{n\to\infty} diam(A_n(x)) \to 0$$

and

$$\lim_{n\to\infty} \frac{|\{j : A_{n,j} \cap S \neq \varnothing\}|\log n}{n} = 0,$$

then the strong consistency of

$$m'_n = \begin{cases} \dfrac{\sum_{i=1}^n Y_i \mathcal{I}(X_i \in A_n(x))}{\sum_{i=1}^n \mathcal{I}(X_i \in A_n(x))}, & \sum_{i=1}^n \mathcal{I}(X_i \in A_n(x)) > \log n \\ 0, \text{其他} \end{cases}$$

is obtained. Since the sample number of each cell is at least $k_n$, i.e.,

$$N(A_n(x)) = \sum_{i=1}^n \mathcal{I}(X_i \in A_n(x)) \geq k_n.$$

From $k_n / \log n \to \infty (n \to \infty)$, when $n$ is sufficiently large,

$$\sum_{i=1}^n \mathcal{I}(X_i \in A_n(x)) \geq k_n > \log n.$$

In this case

$$m'_n = \hat{y}(x) = \frac{\sum_{i=1}^n Y_i \mathcal{I}(X_i \in A_n(x))}{\sum_{i=1}^n \mathcal{I}(X_i \in A_n(x))} = \frac{1}{N(A_n(x))} \sum_{(X,Y)\in A_n(x)} Y.$$

That is, the base regression tree is universally strongly consistent. Therefore, we only need to prove that the conditions of the above two limits are true. The former condition has been proved in the consistency proof of classification DMRF algorithm, so only the latter is needed to prove.

For the base tree with $n$ training samples, there is at most $\dfrac{n}{k_n}$ split regions,

$$\frac{|\{j: A_{n,j} \cap S \neq \varnothing\}| \log n}{n} \leq \frac{n}{k_n} \cdot \frac{\log n}{n} = \frac{\log n}{k_n} \to 0 (n \to \infty).$$

The latter is true. Therefore, the strong consistency of the regression DMRF algorithm is obtained.

□

**Publisher's Note**